\documentclass[conference]{IEEEtran}
\IEEEoverridecommandlockouts
\usepackage{cite}
\usepackage{amsmath,amssymb,amsfonts}
\usepackage{algorithmic}
\usepackage{graphicx}
\usepackage{textcomp}
\usepackage{xcolor}
\usepackage{url}

\begin{document}

\title{Selective Fine-Tuning of GPT Architectures for Parameter-Efficient Clinical Text Classification

}

\author{
\IEEEauthorblockN{1\textsuperscript{st} Fariba Afrin Irany}
\IEEEauthorblockA{
Clarksville, TN, USA \\
afrinirany75@gmail.com
}
\and
\IEEEauthorblockN{2\textsuperscript{nd} Sampson Akwafuo}
\IEEEauthorblockA{
\textit{California State University} \\
Fullerton, California, USA \\
sakwafuo@acm.org
}
}


\maketitle

\begin{abstract}
The rapid expansion of electronic health record (EHR) systems has generated large volumes of unstructured clinical narratives that contain valuable information for disease identification, patient cohort discovery, and clinical decision support. Extracting structured knowledge from these free-text documents remains challenging because clinical language is highly specialized, labeled datasets are limited, and full fine-tuning of large pretrained language models can require substantial computational resources. Efficient adaptation strategies are therefore essential for practical clinical natural language processing applications.

This study proposes a parameter-efficient selective fine-tuning framework for adapting GPT-2 to clinical text classification tasks. Instead of updating the entire pretrained model, the majority of network parameters are frozen, and only the final Transformer block, the final layer normalization module, and a lightweight classification head are updated during training. This design substantially reduces the number of trainable parameters while preserving the contextual representation capabilities learned during pretraining.

The proposed approach is evaluated using radiology reports from the MIMIC-IV-Note dataset with automatically derived CheXpert-style labels. Experiments on 50,000 radiology reports demonstrate that selective fine-tuning achieves approximately 91\% classification accuracy while updating fewer than 6\% of the model parameters. Comparative experiments with head-only training and full-model fine-tuning show that the proposed method provides a favorable balance between predictive performance and computational efficiency. These results indicate that selective fine-tuning offers an efficient and scalable framework for clinical text classification.
\end{abstract}

\begin{IEEEkeywords}
Clinical NLP, Electronic Health Records, GPT-2, Radiology Report Classification, Parameter-Efficient Learning, Transformer Architecture
\end{IEEEkeywords}

\section{Introduction}

The widespread adoption of electronic health record (EHR) systems has generated a large volume of unstructured clinical text, including progress notes, discharge summaries, operative reports, and radiology reports. These narratives contain rich descriptions of patient conditions, diagnostic reasoning, and treatment decisions that are often not captured in structured database fields. As a result, automated analysis of clinical text has become increasingly important for tasks such as disease phenotyping, cohort identification, outcome prediction, healthcare quality assessment, and clinical decision support.

However, clinical narratives present several challenges for computational analysis. Clinical language is highly specialized and frequently contains abbreviations, incomplete sentences, and institution-specific terminology. Early clinical natural language processing (NLP) systems relied primarily on rule-based methods and manually engineered features. Although such approaches can perform well in limited scenarios, they are difficult to scale and often fail to capture the complex contextual relationships present in long clinical documents.

Recent advances in deep learning have significantly improved NLP capabilities. In particular, Transformer-based architectures have become the dominant paradigm for modeling sequential text data. Large pretrained language models learn contextual representations from massive text corpora and can be adapted to downstream tasks through transfer learning. Generative Pretrained Transformer models such as GPT-2 capture syntactic and semantic relationships across long text sequences using autoregressive language modeling \cite{radford2019language}. 

Despite these advances, applying large pretrained models to clinical text introduces practical challenges. Clinical datasets are typically much smaller than the datasets used for model pretraining, which increases the risk of overfitting during full-model fine-tuning. In addition, updating all parameters of large language models requires substantial computational resources and memory. These limitations can make full-model fine-tuning inefficient in many academic and healthcare research environments.

To address these challenges, this study investigates a parameter-efficient selective fine-tuning strategy for adapting GPT-based models to clinical text classification. Instead of updating the entire model, most pretrained parameters are frozen while only the final Transformer block, the final layer normalization layer, and a lightweight classification head are optimized. This approach leverages the hierarchical representation structure of Transformer models while substantially reducing training complexity.

The proposed framework is evaluated on radiology reports from the MIMIC-IV clinical database using automatically derived CheXpert-style labels. Experimental results demonstrate that selective fine-tuning achieves competitive classification performance while updating only a small fraction of model parameters, making it a practical approach for clinical NLP applications under limited computational resources.

\section{Related Work}

Recent advances in natural language processing have been driven by the development of Transformer-based architectures and large-scale pretrained language models. The Transformer architecture introduced the self-attention mechanism, enabling efficient modeling of long-range dependencies and parallel sequence processing \cite{vaswani2017attention}. Subsequent research improved the training stability and optimization behavior of Transformer models through refinements in normalization strategies and architectural design \cite{xiong2020layer}. These developments enabled the emergence of large pretrained language models trained using self-supervised objectives that learn contextual linguistic representations transferable to a wide range of downstream tasks.

Early work on contextual representation learning demonstrated the effectiveness of pretrained embeddings for improving performance in natural language processing tasks. Models such as ELMo \cite{peters2018elmo} and ULMFiT \cite{howard2018universal} showed that pretrained language representations significantly improve downstream performance when labeled datasets are limited. Later architectures expanded both the scale and capability of pretrained models. Generative pretrained Transformer models introduced autoregressive language modeling at scale \cite{radford2018improving,radford2019language,brown2020language}, while bidirectional models such as BERT demonstrated the effectiveness of masked language modeling for capturing contextual relationships across sequences \cite{devlin2019bert}. Several extensions including RoBERTa, XLNet, ALBERT, and T5 further improved model scalability, training efficiency, and transfer learning performance \cite{liu2019roberta,lan2019albert,yang2019xlnet,raffel2020t5}. These models established a general paradigm in which large pretrained models are adapted to downstream tasks through fine-tuning.

Transfer learning has therefore become a central paradigm in modern NLP because pretrained models can be adapted to specialized tasks using relatively small labeled datasets. However, domain mismatch between general-language training corpora and specialized domains often requires additional adaptation strategies. Domain-adaptive pretraining and task-adaptive pretraining approaches have been shown to significantly improve model performance in domain-specific settings \cite{peters2019tune}. These strategies are particularly important in biomedical and clinical natural language processing, where specialized terminology and writing conventions differ substantially from general-language corpora. As a result, pretrained language models are increasingly applied to clinical text classification, information extraction, and healthcare prediction tasks using electronic health record data.

In parallel with advances in language modeling, computational frameworks have been developed to analyze large-scale healthcare, epidemiological, and complex network datasets. Data-driven modeling techniques have been used to analyze disease transmission patterns, epidemiological trends, and health system data integration using large heterogeneous datasets \cite{irany2024large,irany2024bias}. Network-based analytical methods have also been applied to identify structural patterns and community relationships in multilayer data environments \cite{santra2023efficient}. Optimization and simulation approaches have been explored for improving healthcare logistics, emergency response planning, and resource allocation in complex systems \cite{Sakwafuo2020optimization}. Mathematical modeling approaches have also been applied to estimate infectious disease transmission dynamics and public health risk using computational epidemiology frameworks \cite{irany2020estimating}. These studies highlight the growing role of scalable computational modeling and data-driven analytics in modern healthcare research.

Despite their strong performance, applying large language models in clinical settings is challenging due to limited labeled data and high computational costs for full fine-tuning. Parameter-efficient methods such as adapters and low-rank adaptation address this issue by updating only a small subset of parameters while keeping most pretrained weights fixed \cite{houlsby2019adapters,hu2022lora}.



\section{Methodology}

Experiments use the GPT-2 (small) architecture, which contains 12 Transformer blocks, a hidden size of 768, and about 124 million parameters \cite{radford2019language}. The pretrained model is initialized with publicly available weights and adapted for clinical text classification. Using the same pretrained backbone ensures a fair comparison across training strategies.

Three fine-tuning strategies are evaluated to study the trade-off between computational efficiency and predictive performance. In \textit{linear head-only} training, all Transformer parameters remain frozen and only the classification head is trained, which tests the transferability of pretrained representations \cite{peters2019tune}. In \textit{full fine-tuning}, all parameters—including embeddings, Transformer layers, normalization layers, and the classifier—are updated during training \cite{devlin2019bert,radford2019language}.

\begin{figure*}[!htbp]
\centering
\includegraphics[height=12cm,width=12cm]{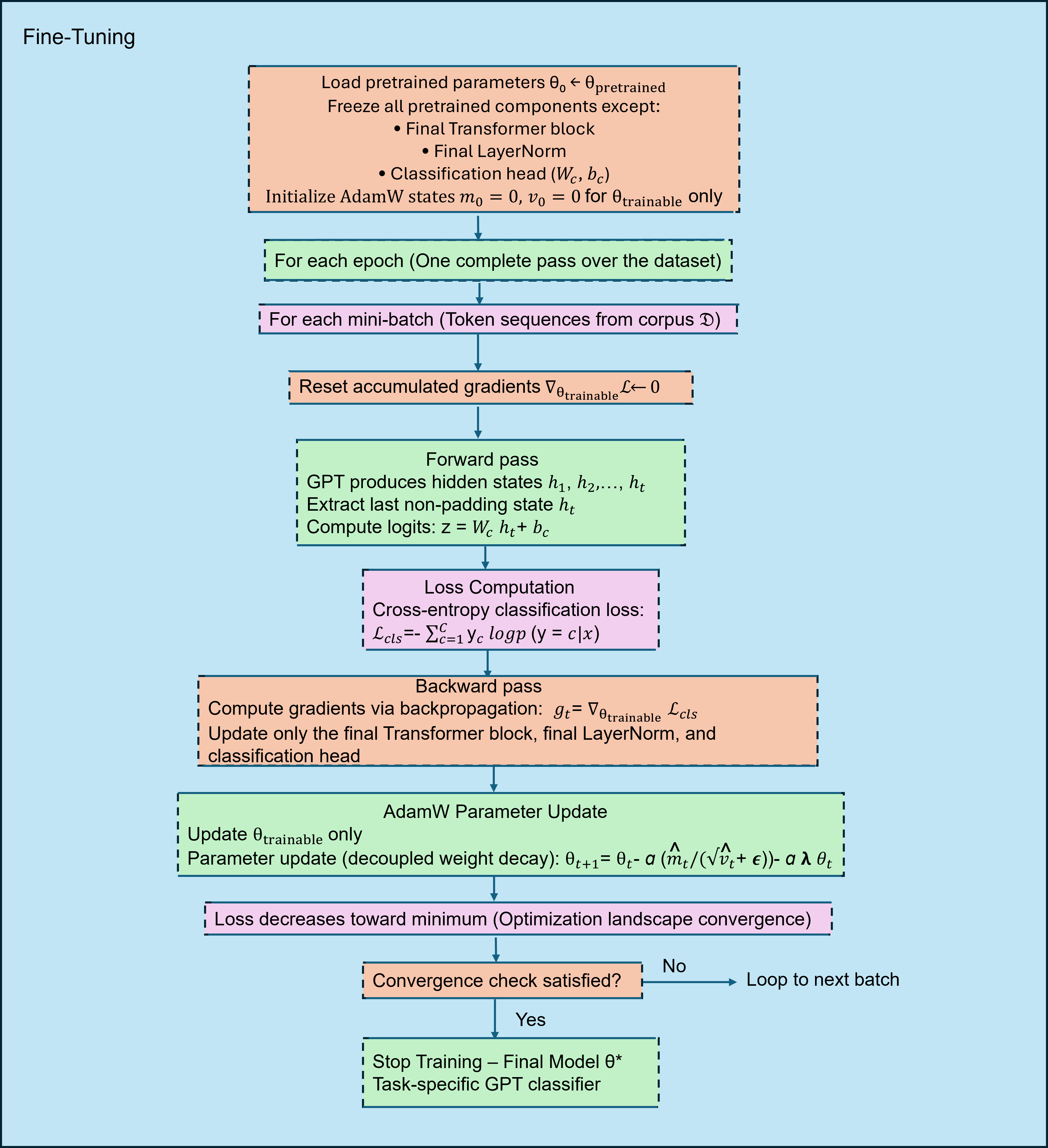}
\caption{Selective fine-tuning workflow for a GPT-based classification model. Pretrained parameters are loaded and all layers are frozen except the final Transformer block, final LayerNorm, and the task-specific classification head. Fine-tuning proceeds over mini-batches using cross-entropy loss and AdamW updates applied only to trainable parameters.}
\label{fine-tuning}
\end{figure*}

The proposed \textit{selective fine-tuning} strategy updates only the final Transformer block, the final layer normalization, and the classification head while keeping the remaining layers frozen. Since higher Transformer layers capture more task-specific representations, this approach enables domain adaptation to clinical text while preserving general linguistic knowledge. The overall selective fine-tuning workflow is illustrated in Figure~\ref{fine-tuning}.

\subsection{Problem Formulation}

Given a labeled dataset
\begin{equation}
\mathcal{D}=\{(x^{(i)},y^{(i)})\}_{i=1}^{N},
\end{equation}
each input sequence $x^{(i)}=(x_1,x_2,\dots,x_T)$ represents tokenized clinical text, and $y^{(i)}$ denotes the corresponding class label. The objective is to learn a classifier
\begin{equation}
f_\theta(x)\rightarrow y
\end{equation}
where $f_\theta$ is a GPT-based neural network parameterized by $\theta$.

\subsection{Model Initialization}
Fine-tuning begins by initializing model parameters from a pretrained GPT checkpoint
\begin{equation}
\theta_0 \leftarrow \theta_{\text{pretrained}}
\end{equation}
which transfers linguistic knowledge learned from large-scale self-supervised training to the downstream clinical classification task \cite{radford2019language,vaswani2017attention}.

\subsection{Input Representation}
Clinical text sequences are tokenized using the GPT tokenizer and converted to a fixed-length representation through truncation and padding. For a mini-batch of size $B$, the data loader produces
\begin{equation}
\mathbf{X}\in\mathbb{R}^{B\times T}, \quad
\mathbf{M}\in\{0,1\}^{B\times T}, \quad
\mathbf{y}\in\mathbb{R}^{B}
\end{equation}
where $\mathbf{X}$ contains token indices, $\mathbf{M}$ is an attention mask, and $\mathbf{y}$ contains ground-truth labels \cite{paszke2019pytorch}.

\subsection{Selective Fine-Tuning}
To improve efficiency, most pretrained parameters are frozen during training. Only the following components remain trainable:

\begin{itemize}
\item the final Transformer block
\item the final layer normalization module
\item the task-specific classification head
\end{itemize}

Let
\begin{equation}
\theta=\{\theta_{\text{frozen}},\theta_{\text{trainable}}\}
\end{equation}
where
\begin{equation}
\theta_{\text{trainable}}=
\{\theta_{\text{block-}L},\theta_{\text{LN-final}},W_c,b_c\}.
\end{equation}
Higher Transformer layers capture more task-specific representations, while lower layers encode general linguistic knowledge \cite{peters2019tune}.

\subsection{Classification Head}
A linear classification head is attached to the final sequence representation. For $C$ classes,
\begin{equation}
\mathbf{z}=W_c h_T+b_c
\end{equation}
where $W_c\in\mathbb{R}^{C\times d_{\text{model}}}$ and $b_c\in\mathbb{R}^{C}$ are trainable parameters. Since GPT is a decoder-only architecture without a classification token, the hidden state corresponding to the last non-padding token $h_T$ is used as the sequence representation \cite{radford2019language}.

\subsection{Training and Optimization}
Training proceeds for a fixed number of epochs using mini-batch stochastic optimization. Gradients are reset before each update
\begin{equation}
\nabla\theta_{\text{trainable}} \leftarrow 0
\end{equation}
and the model produces hidden states
\begin{equation}
\mathbf{H}=(h_1,h_2,\dots,h_T)
\end{equation}
from which the final representation $h_T$ is extracted.
Class probabilities are computed using softmax
\begin{equation}
p(y=c|x)=\frac{\exp(z_c)}{\sum_{j=1}^{C}\exp(z_j)}
\end{equation}
and the model is trained using cross-entropy loss
\begin{equation}
\mathcal{L}_{cls}
=
-\sum_{c=1}^{C} y_c \log p(y=c|x).
\end{equation}
Parameters in $\theta_{\text{trainable}}$ are updated using the AdamW optimizer
\begin{equation}
\theta_{\text{trainable}}
\leftarrow
\theta_{\text{trainable}}
-\alpha \, \text{AdamW}(g_t)
\end{equation}
where $\alpha$ denotes the learning rate \cite{loshchilov2019adamw}.

\subsection{Performance Evaluation}
Model performance is evaluated using classification accuracy
\begin{equation}
\text{Accuracy}
=
\frac{1}{N}\sum_{i=1}^{N}\mathbb{I}(\hat{y}_i=y_i)
\end{equation}
where
\begin{equation}
\hat{y}_i=\arg\max_c z_{i,c}.
\end{equation}
Training proceeds for a predefined number of epochs, producing the final task-adapted model
\begin{equation}
\theta^{*}=\arg\min_{\theta_{\text{trainable}}}\mathcal{L}_{cls}.
\end{equation}

\subsection{Training Configuration}
Model training was performed using mini-batch stochastic optimization with the AdamW optimizer and a learning rate of $2\times10^{-5}$. 
The batch size was set to 16 and all models were trained for 10 epochs. Clinical text documents were tokenized using the GPT tokenizer and truncated to a maximum sequence length of 512 tokens to ensure consistent input dimensions.

For classification, the contextual representation produced by the Transformer decoder stack was passed to a task-specific classification head, and model parameters were optimized using the cross-entropy loss function.  


Experiments were conducted using PyTorch and the HuggingFace Transformers library on an NVIDIA GPU(A100).

\section{Selective Fine-Tuning Parameterization and Time Complexity Analysis}

The proposed model adopts a selective fine-tuning strategy based on GPT-2 (small) \cite{radford2019language}. Instead of updating all parameters, most pretrained components are frozen and only the top layers are optimized for the downstream classification task. Specifically, the embedding layers and the first $11$ Transformer blocks are kept fixed, while the final Transformer block, the final layer normalization layer, and the task-specific classification head remain trainable.

GPT-2 (small) contains approximately $124$ million parameters. The embedding layers alone contain about $38.6$ million parameters, while each Transformer block contains approximately $7.08$ million parameters. Under the proposed setup, only the final Transformer block and lightweight output layers are updated, resulting in roughly $7.08$ million trainable parameters. This corresponds to updating less than $6\%$ of the total model parameters.

From a computational perspective, the dominant cost arises from the final Transformer block. For sequence length $T$, the self-attention operation scales as $\mathcal{O}(T^2 d_{\text{model}})$, while the feed-forward network scales as $\mathcal{O}(T d_{\text{model}} d_{\text{ff}})$ \cite{vaswani2017attention}. 

\begin{figure*}[!htbp]
    \centering
       \includegraphics[height=5cm, width=10cm]{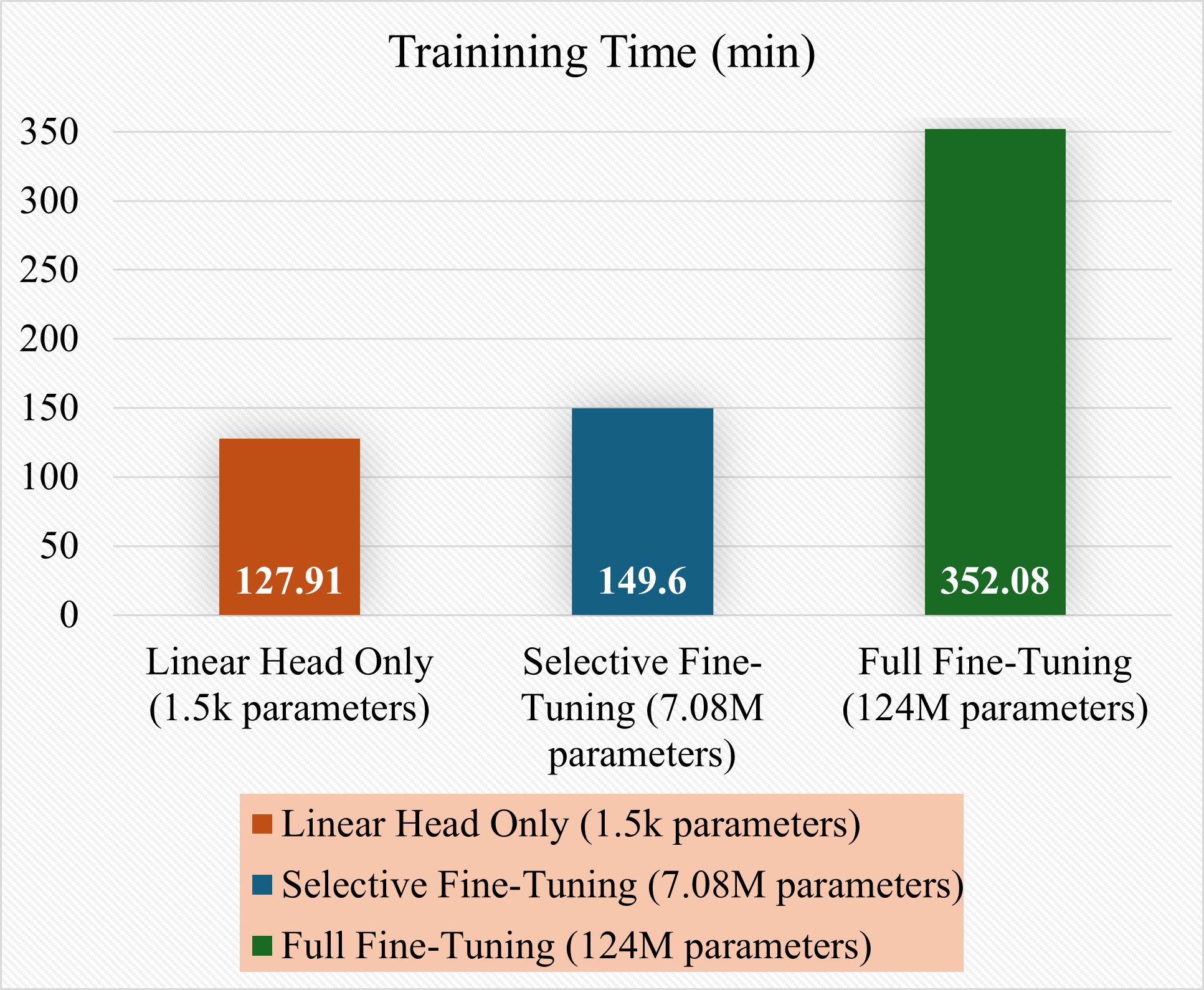} %
    \caption{Total training time for different fine-tuning strategies.}
    \label{fig:training_time}
\end{figure*}

\begin{figure*}[!htbp]
    \centering
    \includegraphics[height=5cm, width=10cm]{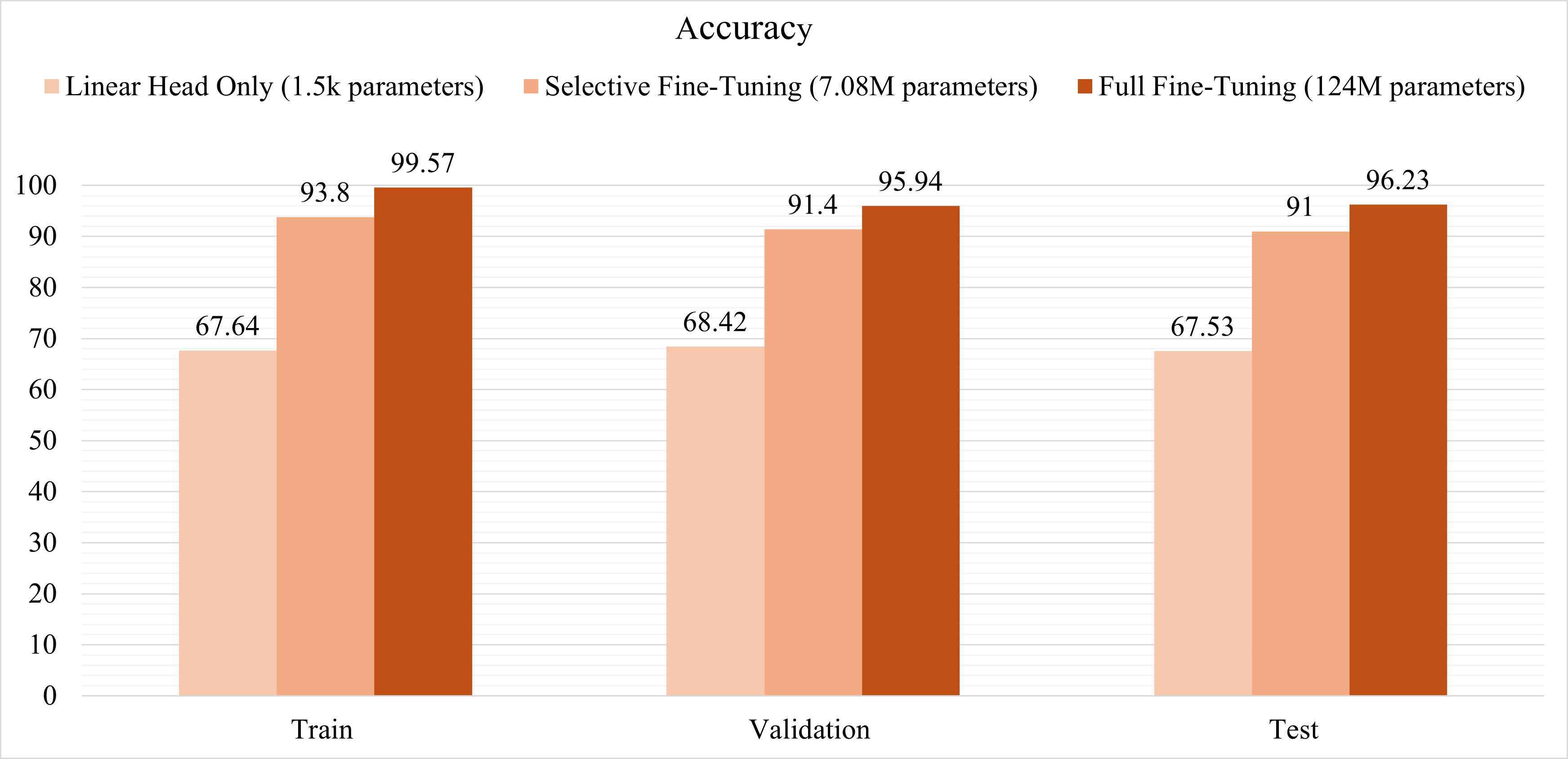} %
    \caption{Classification accuracy on training, validation, and test sets across fine-tuning strategies.}
    \label{fig:accuracy}
\end{figure*}

\begin{figure*}[!htbp]
    \centering
   \includegraphics[height=5cm, width=10cm]{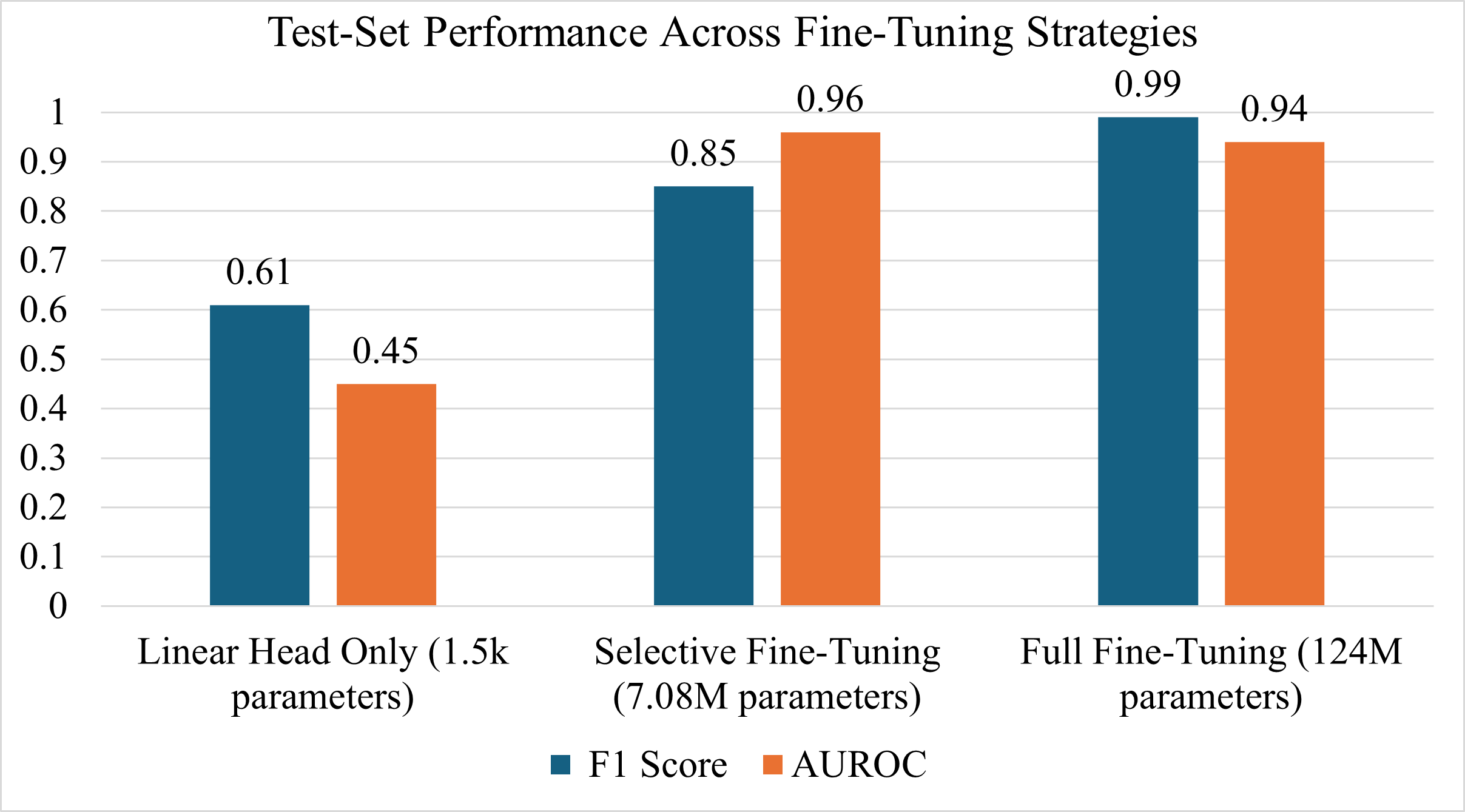}
    \caption{Test-set performance across fine-tuning strategies. The figure compares F1 score and AUROC for linear head–only training, selective fine-tuning of final transformer block, and full model fine-tuning, illustrating the trade-off between predictive performance and the number of trainable parameters.}
    \label{fig:test_set_performance}
\end{figure*}

\section{Dataset and Label Construction}
\begin{table}[h]
\centering
\caption{Label Distribution for Binary Classification Task}
\begin{tabular}{lcc}
\hline
\textbf{Label Category} & \textbf{Number of Reports} & \textbf{Percentage} \\
\hline
Positive or Uncertain (1) & 16,501 & 33.0\% \\
Negative (0) & 33,499 & 67.0\% \\
\hline
\textbf{Total} & \textbf{50,000} & \textbf{100\%} \\
\hline
\end{tabular}
\label{tab:label_distributionv}
\end{table}

Experiments were conducted using radiology reports from MIMIC-IV-Note v2.2, a large de-identified clinical text dataset available through PhysioNet \cite{johnson2023mimiciv}. A subset of reports was used to construct a labeled dataset for supervised learning. The dataset used in this study consists of 50,000 radiology reports extracted from the MIMIC-IV clinical database. Each report was automatically labeled using a CheXpert-style rule-based extraction approach. For the purposes of this study, the classification task was formulated as a binary problem in which reports containing either positive or uncertain findings were grouped into a single class, while reports with no evidence of disease were assigned to the negative class. This formulation follows common practice in clinical text classification where uncertain findings are treated as clinically relevant indicators. Table ~\ref{tab:label_distributionv} summarizes the label distribution for the binary classification task.

Because structured diagnosis labels were not consistently available for all reports, CheXpert-style weak labels were derived directly from report text using rule-based extraction with negation and uncertainty handling \cite{irvin2019chexpert}. For each report, 14 condition labels were generated and stored together with the original identifiers and report text in a structured BigQuery table for downstream GPT-based classification experiments. This labeling strategy enables scalable supervision from real-world clinical narratives while preserving clinically meaningful uncertainty. 

\begin{table*}[!htbp]
\centering
\caption{Comparison of model performance and training efficiency across fine-tuning strategies. Selective fine-tuning achieves strong generalization performance while substantially reducing the number of trainable parameters and training time per epoch compared to full fine-tuning.}
\label{tab:model_comparison}
\begin{tabular}{lcccccc}

Fine-Tuning Strategy & Trainable Parameters & Time / Epoch (min) & Val. Acc. & Test Acc. & F1 Score & AUROC \\

Linear head only & 1.5k & 12.79 & 68.42 & 67.53 & 0.61 & 0.45 \\
Selective fine-tuning (proposed) & 7.08M & 14.96 & 91.40 & 91.00 & 0.85 & 0.96 \\
Full fine-tuning & 124M & 35.21 & 95.94 & 96.23 & 0.99 & 0.94 \\

\end{tabular}
\end{table*}

\section{Results}
This section presents the experimental evaluation of the proposed selective fine-tuning strategy for clinical text classification using GPT-2. 
The experiments compare three fine-tuning approaches applied to the same pretrained GPT-2 backbone: 
(i) linear classification head–only training, 
(ii) selective fine-tuning of the final Transformer block (proposed method), and 
(iii) full model fine-tuning. 

The comparison focuses on both predictive performance and computational efficiency for the binary classification task \texttt{label\_any\_disease\_pos\_or\_unc} using 50{,}000 radiology reports from the MIMIC-IV dataset. 
To ensure unbiased evaluation, the dataset was randomly divided into training (70\%), validation (15\%), and test (15\%) subsets.

Table~\ref{tab:model_comparison} summarizes the performance and training efficiency of the three strategies. Linear head–only training involves updating only the final classification layer while freezing the transformer backbone, resulting in the smallest number of trainable parameters (1.5k). Although this approach requires the lowest computational cost, it achieves limited predictive performance, with a validation accuracy of 68.42\% and a test accuracy of 67.53\%. The corresponding F1 score and AUROC values are also comparatively low, indicating that the model cannot effectively adapt pretrained representations to the clinical text domain when the backbone remains frozen. 




Figure~\ref{fig:training_time} compares the total training time required for each strategy. Linear head–only training completes the fastest due to the small number of trainable parameters, while selective fine-tuning requires slightly more training time. In contrast, full fine-tuning requires significantly longer training time because gradients must be computed for all transformer layers.

Figure~\ref{fig:accuracy} shows classification accuracy across the training, validation, and test datasets. Head-only training consistently produces the lowest accuracy across all splits. Selective fine-tuning yields a large improvement in performance while maintaining stable generalization between validation and test sets. Although full fine-tuning achieves the highest training accuracy, the improvement over selective fine-tuning on the validation and test sets is comparatively modest.

Figure~\ref{fig:test_set_performance} further compares the F1 score and AUROC obtained on the test set. The results indicate that selective fine-tuning achieves performance close to full fine-tuning while requiring substantially fewer trainable parameters.

\section{Data Availability and Reproducibility}
This research used the MIMIC-IV clinical database, which is accessible to credentialed researchers through PhysioNet. In accordance with the data use agreement, patient-level clinical text and derived datasets are not shared by the authors. Reproducibility is supported through publicly available code and Google Colab notebooks at:
\url{https://drive.google.com/drive/folders/1GI21lZZiRAiI8Y-C5FEU1PsTklN0487X}



\section{Conclusion}

This study investigates adapting a pretrained GPT-2 model for clinical text classification using a parameter-efficient selective fine-tuning strategy. Instead of updating the entire model, most pretrained parameters remain frozen while only the final Transformer block, the final layer normalization layer, and a lightweight classification head are trained. This approach reduces the number of trainable parameters while preserving contextual knowledge learned during pretraining.

Experiments on 50,000 radiology reports from the MIMIC-IV-Note dataset show that selective fine-tuning achieves strong classification performance with significantly lower computational cost. Comparisons with head-only training and full-model fine-tuning illustrate the trade-off between efficiency and predictive accuracy, where selective fine-tuning maintains competitive performance while updating only a small portion of model parameters.

A limitation of this study is that experiments were conducted only on radiology reports from the MIMIC-IV dataset. Future work will evaluate other clinical datasets and compare additional parameter-efficient techniques such as adapters and low-rank adaptation methods.

\end{document}